\documentclass{article}

\usepackage{spconf}

\usepackage[utf8]{inputenc} % allow utf-8 input
\usepackage[T1]{fontenc}    % use 8-bit T1 fonts
\usepackage{hyperref}       % hyperlinks
\usepackage{url}            % simple URL typesetting
\usepackage{booktabs}       % professional-quality tables
\usepackage{amsfonts}       % blackboard math symbols
\usepackage{nicefrac}       % compact symbols for 1/2, etc.
\usepackage{microtype}      % microtypography
\usepackage{lipsum}
\usepackage{graphicx}

\usepackage{amsmath}
\usepackage{cite}

\title{PointVoteNet: Accurate Object Detection and\\6 DoF Pose Estimation in Point Clouds}
\name{Frederik Hagelskj{\ae}r, Anders Glent Buch}
\address{SDU Robotics, University of Southern Denmark\\
\texttt{\{frhag,anbu\}@mmmi.sdu.dk}}

\begin{document}
\maketitle

\begin{abstract}
We present a learning-based method for 6 DoF pose estimation of rigid objects in point cloud data. Many recent learning-based approaches use primarily RGB information for detecting objects, in some cases with an added refinement step using depth data. Our method consumes unordered point sets with/without RGB information, from initial detection to the final transformation estimation stage. This allows us to achieve accurate pose estimates, in some cases surpassing state of the art methods trained on the same data.
\end{abstract}

\keywords{Deep learning, Point clouds, Pose estimation}

\section{Introduction}
Deep learning has reigned supreme for almost a decade now, with applications in a number of domains, perhaps most notably computer vision. A key challenge for visual processing systems is detection and accurate localization of rigid objects, allowing for a number of applications, including robotic manipulation. The most mature branch of deep learning is arguably CNN-based models, which are now applied in a variety of contexts. For object detection, CNNs are currently providing the best building blocks for learning-based detection systems in 2D, where the relevant objects are localized by a bounding box. More recently, full 6 DoF pose estimation systems have been demonstrated using similar techniques.

Some methods rely on RGB inputs only to reconstruct the full 6 DoF object pose. In \cite{brachmann2016uncertainty} an uncertainty-aware regression method provides dense predictions of 3D object coordinates in the input image. Only keypoints are detected in \cite{rad2017bb8} and \cite{tekin2018real}, in the latter case using a YOLO-like \cite{redmon2016you} backbone. In \cite{xiang2017posecnn} a prediction is made for the object center, followed by a regression towards the rotation. The method in \cite{peng2019pvnet} uses a compromising approach where a semi-dense set of keypoints are predicted over the image.

Methods relying on RGB-D data often augment an initial RGB-based detection with a refinement step using the depth information. \cite{brachmann2014learning} again casts dense pixel-wise votes for 3D object coordinates, but using RGB-D inputs. In \cite{kehl2017ssd}, an SSD-like detector \cite{liu2016ssd} provides initial detections, which are refined using depth information. Initial segmentations are used in \cite{wang2019densefusion} and finally the pose is refined using PointNet \cite{qi2017pointnet}.

Some of these works use point cloud data, but not in the global detection phase, rather as a component during refinement. Our work, on the other hand, uses raw, unordered 3D point cloud data from end to end, i.e. both during initial detection of candidate regions of the scene and during actual transformation estimation and refinement. Our method relies on a PointNet backbone \cite{qi2017pointnet} to achieve this, and we show that in many of the tested circumstances, this significantly increases detection rates and localization accuracy during full 6 DoF pose estimation of objects.

\begin{figure}[ht]
    \begin{center}
        \includegraphics[width=0.70\linewidth]{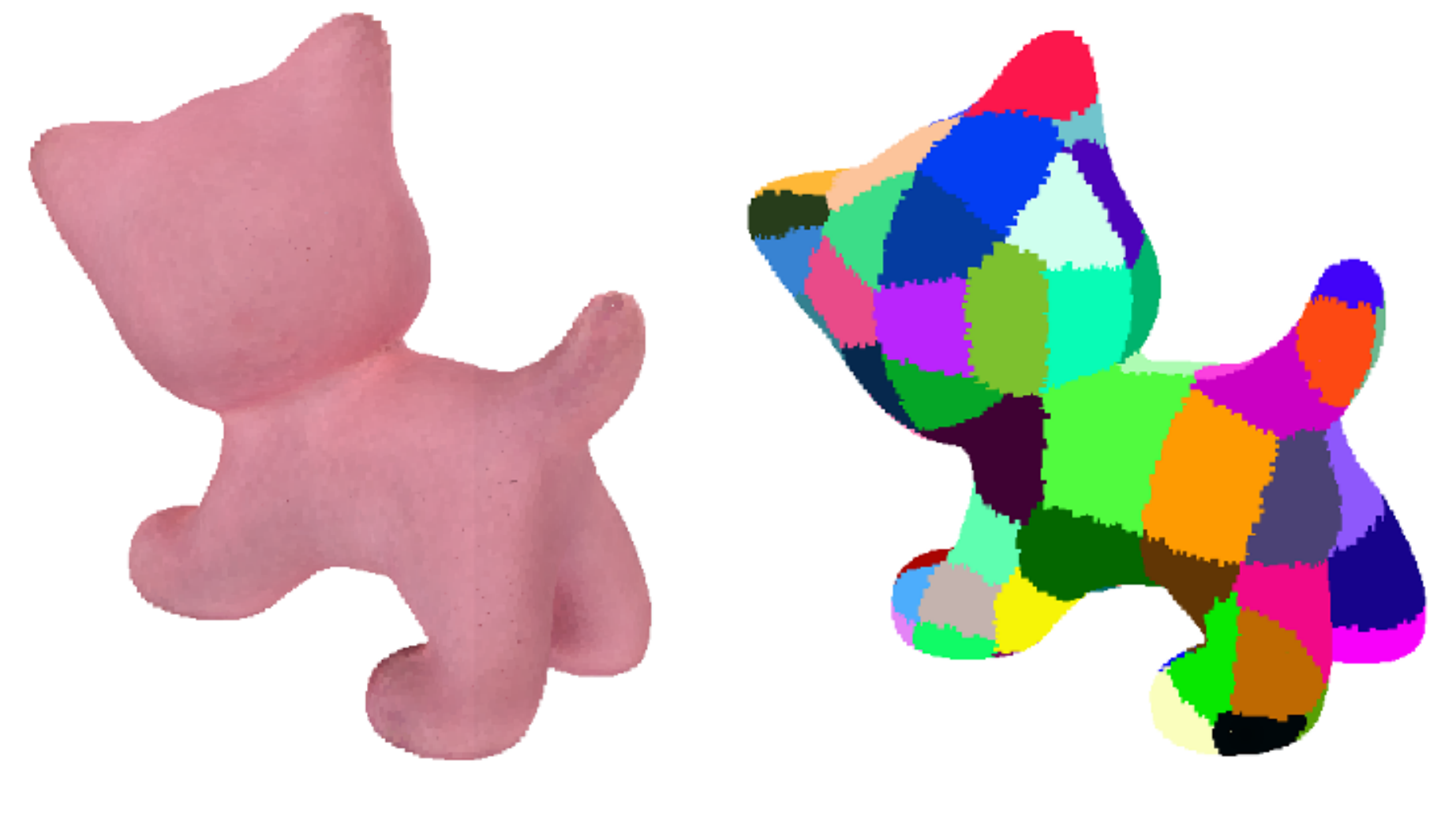}
        \caption{Object model and resulting segments for each point.}
        \label{fig:segments}
        \vspace{-10pt}
    \end{center}
\end{figure}

We describe here a method for training a joint classification and segmentation to directly propose detections and perform point-to-point correspondence assignments between an observed scene point cloud and a known 3D object model. An example of these segments of an object is shown in Fig.~\ref{fig:segments}. These correspondences are passed to a voting algorithm that performs density estimates directly in the 6-dimensional pose parameter space. We also show how the addition of color coordinates to the input point cloud dramatically increases the accuracy of our method.

\section{Method}
This section gives a full description of our method. We start with a description of all the processing stages of our pipeline during inference, followed by a detailed description of the training process.

\begin{figure*}[ht]
    \begin{center}
        \includegraphics[width=0.99\linewidth]{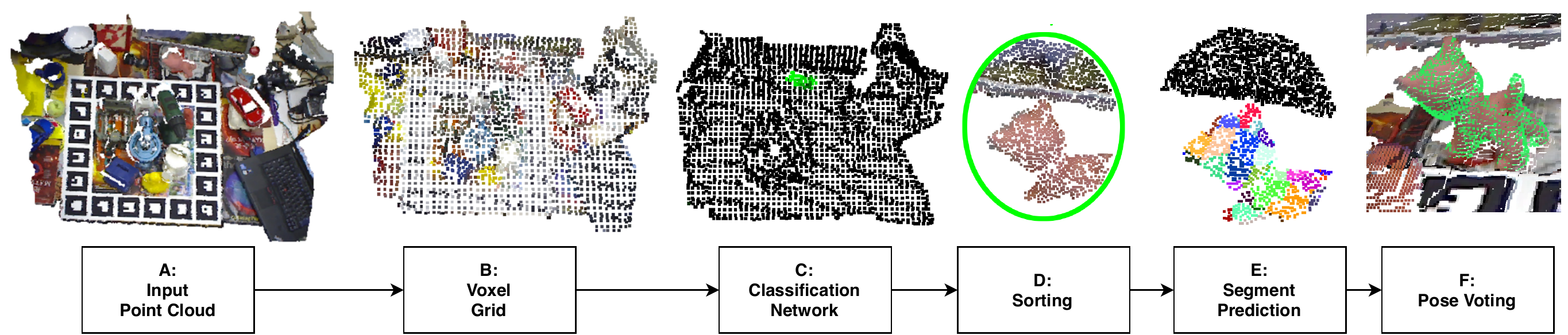}
        \caption{Visualization of the pose estimation process. First, the original point cloud (A) is sub-sampled into anchor-points (B) using a voxel-grid. For each anchor-point, a spherical point cloud is collected. Using PointNet, a score is returned (C) showing the likelihood of the object being present (green). The responses are then sorted, and the 16 highest scoring spheres returned (D). The PointNet segmentation part is then used to create votes for each point in the CAD model (E), where black represents background. Finally, these votes are used to find a pose estimation for the object (F).}
        \label{fig:pipeline}
    \end{center}
\end{figure*}

\subsection{Inference pipeline}
\textbf{Object representation:} The goal of our pipeline is to provide the full 6 DoF pose of an object, in our case provided as either a CAD model or as a dense point cloud. To aid the training process, we represent this model by a very low number of keypoints, uniformly sampled on the object surface with an empirically chosen distance of 2.5~cm. A voxel grid achieves this downsampling very efficiently \cite{rusu20113d} and we end up with a number of keypoints between approx. 50 and a few hundreds, depending on the size of the object. In the top part of Fig.~\ref{fig:pipeline} we show an example of a model next to a coloring of the model points according to the nearest keypoint.

\textbf{Input preprocessing:} The input to our system during inference is a novel scene, provided as an unordered point cloud containing XYZ and, when applicable, RGB coordinates. The first stage is a binary classifier, aimed at determining candidate points in the scene where the searched object is located. To limit the data, the scene is uniformly downsampled---again with an empirically chosen 2.5~cm point spacing---to approx. 3000-5000 anchor points, depending on the size of the scene. We show a visualization of this in the middle of Fig.~\ref{fig:pipeline}.

\textbf{Proposal via classification:} The thousands of uniform anchor points are each seen as candidate positions for the center of the searched object. In order to detect promising positions, we use a binary PointNet classifier \cite{qi2017pointnet} to estimate the probability of the presence of the object at each anchor point. More specifically, we sample 2048 scene points in the spherical neighborhood around an anchor point and pass these to a PointNet with a single logistic output neuron. The choice of the number of point neighbors is a trade-off between speed and accuracy, and we have observed a saturation in accuracy around the chosen 2048 points. In the bottom part of Fig.~\ref{fig:pipeline} we show an example of such a neighborhood.

\textbf{Correspondences via segmentation:} The classifier allows to greatly prune the search space by only considering the top predictions from the object-specific classifier. At this stage we process only the most promising anchor points and their neighborhoods. Non-maximum suppression is not performed on neighbouring anchor points, as allowing overlapping predictions improves the performance. The objective now is to associate each of the 2048 points around the anchor to either background or to the corresponding point on the given object surface. If the object is reduced to $K$ keypoints, we now invoke a $(K+1)$-way point-wise semantic segmentation network (again inspired by PointNet) to label each point as either background or one of the $K$ object keypoints. An example of a labeling is shown in the bottom of Fig.~\ref{fig:pipeline}, where black corresponds to a background prediction. At this stage, we perform segmentation only on the 16 top scoring anchor points.

\textbf{6 DoF pose estimation from correspondences:} The outcome of the labeling process is a set of many-to-few correspondences, where, on average we can expect the translation errors of many of the points in the scene to cancel each other out, as they always vote for the point in the center of each of the object segments. The problem of pose estimation has now been reduced to a classical 3D-3D correspondence problem, where we are given a large set of up to 2048 point-to-point correspondences for each of the 16 top scoring anchors. Many algorithms are available for estimating the relative pose between the two point sets in the presence of mismatches. One of the most effective algorithms is the rotational subgroup voting algorithm \cite{buch2017rotational}, which has shown superior performance for this type of problems. We thus directly pass the many non-background labeled points directly to this algorithm and compute the full 6 DoF rigid transformation between the object and the scene. The 16 poses for each of the processed anchors are refined using a coarse-to-fine ICP \cite{besl1992method}.

\textbf{Multi-modal localization loss:} A final pose verification is performed on the 16 pose estimates by a multi-modal loss that determines how well the estimated pose of the object fits with the observed scene data. The voting algorithm in the previous step already produces a density estimate in the parameter space $SE(3)$, which is proportional to the number and quality of correspondences that vote for the final pose. This, however, has proven insufficient for our method, since that score does not intrinsically include the sensor-specific information that we have available, i.e. a viewing direction and in some cases color information. Our pose verification is thus performed by first transforming the object model points into the scene using the estimated pose. Occluded points, i.e. points lying behind the scene data relative to the camera viewing axis, are removed and the remaining points are paired with the closest points in the scene using a $k$-d~tree. We now compute a geometric and, when applicable, a color loss as RMS errors in the Euclidean and the perceptual RGB space:
\begin{align}
    l_\text{geometric} &= \sqrt{\frac1n \sum_{i=1}^n\left(\mathbf{p}_i-\mathbf{p}_{i,NN})\right)^2} \\
    l_\text{color} &= \sqrt{\frac1n \sum_{i=1}^n\left(\mathbf{c}_i-\mathbf{c}_{i,NN})\right)^2}
\end{align}
where $n$ designates the number of remaining points after occlusion removal and the $NN$ subscript means the nearest neighbor scene point for each of the $n$ transformed object points in $\mathbf p$. In colored point clouds, acquired from e.g. RGB-D sensors, each point also has an associated RGB tuple, which in the second equation is denoted $\mathbf c$. The geometric and perceptual losses are combined with the KDE-based score of the voting algorithm \cite{buch2017rotational} to produce the final localization loss:
\begin{equation}
    l_\text{loc} = \frac{{l_\text{geometric} \cdot l_\text{color}}}{s_\text{KDE}}
\end{equation}
This allows us to separate good from bad pose estimates with high specificity. The final output of our algorithm is the detection that minimizes this loss.

\subsection{Training}
Our algorithm is trained with a number of real examples, to which we apply a limited set of augmentations to prevent overfitting. Similar to existing approaches, e.g. \cite{peng2019pvnet,rad2017bb8,tekin2018real}, the training examples are gathered from real scenes, each annotated with one or more ground truth poses of objects. Object models are given as mesh models, which can be with or without color information. The difference from many other works is the use of raw, unordered point cloud data in our work. To extract as much geometry from the surfaces, we include XYZ coordinates, normal vector coordinates, and the local curvature. The normal vectors/curvature are computed using PCL \cite{rusu20113d} with a support radius of 1~cm. When including color, we add three extra RGB components to each surface point.

\textbf{Data preparation:} For a single annotated example, we start by transforming the provided 3D model into the scene using the ground truth pose. All scene points within a distance threshold of 1~cm of the transformed model are treated as foreground points. Points above a distance threshold of 2~cm are considered as background.\footnote{Due to non-negligible inaccuracies of the provided pose annotations, we need a fairly large distance threshold for foreground points. Additional variation in these ground truth pose inaccuracies, result in a band of approx. 1-2~cm where both foreground and background points occur. To avoid an excessive amount of mislabeled points, we discard all points within this band.} For each of the foreground points, we now associate a segmentation label corresponding to the index of the nearest keypoint on the model.

\textbf{Translation-invariant positives:} Next, we sample 20 random foreground points for creating positive training examples for PointNet. Random sampling on the visible object surface, as opposed to only considering e.g. the centroid, makes our algorithm robust to translations, which will inevitably occur when using uniform anchor points during inference. For each sampled visible point we uniformly extract 2048 scene points within a spherical neighborhood of 0.6 times the 3D bounding box diagonal of the object model. This provides us with 20 positive training examples with both a class label for object presence and 2048 point-wise segmentation labels. All training examples are centered by subtracting the centroid of the points in the sphere.

\textbf{Easy and hard negatives:} The naive way of extracting counter-examples for training would be a fully uniform sampling in the rest of the scene. However, significantly higher specificity can be obtained by including a number of hard negatives during training. Thus, the easy negatives are sampled far enough away from the object instance to not include any foreground points. The hard negatives are sampled in the vicinity of the object, and, although some of the points in these neighborhoods are labeled with a non-background segmentation label, the classification label is still set to non-object. We use 20 easy and 10 hard negatives, which, together with the positives, sums to 50 training examples per annotated object instance, each with 2048 unordered 3D points.

\textbf{Augmentation:} To increase the network's ability to generalize to different views, we perform simple augmentation on top of the 50 per-instance training examples as follows. For each of the positives we remove the background points and insert a randomly sampled background from one of the easy negatives.\footnote{For the smaller training set in our experiments, LINEMOD, this particular augmentation is performed three times per positive example.} The positive cloud is randomly translated a small distance, and random segments around object keypoints are also removed from the point cloud. The background cloud is then translated randomly with a uniform distribution scaled by half the object diagonal. Finally, the point cloud is cut so that all points fit within the sphere of 0.6 times the object diagonal. Another 20 positive examples are created, but now where no background points are inserted. Finally, 20 segments of mixed background segments, i.e. easy negatives, are also created to train on random backgrounds. All in all, in addition to the original 50 per-instance training examples, we augment by 60 examples with equal number of positive and negative classification labels. Finally, all training examples, regular and augmented, are jittered by additive, zero-centered Gaussian noise on all geometry/color coordinates with a standard deviation of 0.01, which for the XYZ coordinates (given in mm) translates to a very small displacement.

\textbf{Symmetry handling:} Some objects have two or more rotational symmetries, in which case the exact rotation around the axis of rotation cannot be determined. To handle this, we reduce symmetric models down to the fewest required distinct keypoints during training set generation. An example is a $\infty$-fold symmetric cylinder, which can be described using only a single line of points along the main axis.

\section{Experiments}
In this section we present experimental results on two of the most well-tested datasets for pose estimation in cluttered scenes, LINEMOD \cite{hinterstoisser2012model} and Occlusion \cite{brachmann2014learning}. Both datasets show one or more objects in several hundred scenes. We use the same split (approx. 15/85~\% for train/test) for the LINEMOD dataset as in earlier works, such as \cite{brachmann2016uncertainty,rad2017bb8,tekin2018real,peng2019pvnet}. For the Occlusion dataset, eight of the LINEMOD sequences make up the training examples. In all our experiments, we rely entirely on unordered point cloud data, which are reconstructed using the provided RGB and depth images. As per convention, the Eggbox and Glue objects are treated as 2- and $\infty$-fold symmetric objects, respectively. We evaluate all our results using the ADD metric mandated by the dataset creators \cite{hinterstoisser2012model,brachmann2014learning}.

%\footnote{Some of the competing methods evaluate their results using IoU and a projective localization error \cite{peng2019pvnet}, both of which are restricted to images.}

\textbf{Training parameters:} We jointly train per-object classification and segmentation heads on top of a standard PointNet architecture. The LINEMOD dataset is trained for 80 epochs and the Occlusion training set, being much larger, is trained for 20 epochs. The remaining training parameters are the same, as follows. The Adam optimizer \cite{kingma2014adam} is used for the optimization with a batch size of 16 and a base learning rate 0.001. We also use a momentum of 0.9 to further speed up convergence. The two cross entropies have different scales and we empirically set the relative importance to 0.15 for the classification loss and 0.85 for the segmentation loss during backpropagation.

\subsection{LINEMOD}
In Tab.~\ref{tab:linemod} we compare results for LINEMOD. The first three methods are using either RGB-D or pure image data. DenseFusion \cite{wang2019densefusion} and our method use colored point cloud data (DenseFusion by image crops, ours directly by RGB components attached to each 3D point). Of these methods, our method is able to produce more accurate pose estimates on average.

\begin{table}[ht]
    \centering
    \small
    \begin{tabular}{|c|c|c|c|c|c|}
        \hline
                & \cite{kehl2017ssd} & \cite{rad2017bb8}  & \cite{peng2019pvnet} & \cite{wang2019densefusion} & Ours\\
        \hline
        Ape     & 65.0 & 40.4 & 43.6 & 92.0 & 80.7 \\
        Bench v. & 80.0 & 91.8 & 99.9 & 93.0 & 100 \\
        Camera & 78.0 & 55.7 & 86.9 & 94.0 & 100 \\
        Can & 86.0 & 64.1 & 95.5 & 93.0 & 99.7 \\
        Cat & 70.0 & 62.6 & 79.3 & 97.0 & 99.8 \\
        Driller & 73.0 & 74.4 & 96.4 & 87.0 & 99.9 \\
        Duck & 66.0 & 44.3 & 52.6 & 92.0 & 97.9 \\
        Eggbox & 100 & 57.8 & 99.2 & 100 & 99.9 \\
        Glue & 100 & 41.2 & 95.7 & 100 & 84.4 \\
        Hole p. & 49.0 & 67.2 & 81.9 & 92.0 & 92.8 \\
        Iron & 78.0 & 84.7 & 98.9 & 97.0 &  100 \\
        Lamp & 73.0 & 76.5 & 99.3 & 95.0 & 100 \\
        Phone & 79.0 & 54.0 & 92.4 & 93.0 & 96.2 \\
        \hline
        Average & 79.0   & 62.7 & 86.3 & 94.3 & 96.3 \\
        \hline
    \end{tabular}
    \caption{LINEMOD results. The competing methods are SSD-6D \cite{kehl2017ssd}, BB8 \cite{rad2017bb8}, PVNet \cite{peng2019pvnet}, and DenseFusion \cite{wang2019densefusion}.}
    \label{tab:linemod}
\end{table}

\subsection{Occlusion}
We show Occlusion dataset results in Tab.~\ref{tab:occlusion_color}. In the first case, we used only XYZ coordinates as inputs to our method. Both PoseCNN \cite{xiang2017posecnn} and PVNet \cite{peng2019pvnet}, which both use image data, produce much less accurate poses for this dataset. On the contrary, when adding a projective ICP refinement step to PoseCNN, this method achieves slightly better results than ours. This is likely due to a) the more sophisticated ICP, compared to the standard 3D point to point ICP used by us, and b) the use of 80000 extra synthetic images during training.

\begin{table}[htb]
    \centering
    \small
    \begin{tabular}{|c|c|c|c||c|c|}
        \hline
                    & \cite{xiang2017posecnn} &  \cite{peng2019pvnet} & Ours & \cite{xiang2017posecnn} & Ours \\
        \hline
        Ape         & 9.60 & 15.0 & 66.2 & 76.2 & 70.0 \\
        Can         & 45.2 & 63.0 & 90.3 & 87.4 & 95.5 \\
        Cat         & 0.93 & 16.0 & 34.7 & 52.2 & 60.8 \\
        Driller     & 41.4 & 25.0 & 59.6 & 90.3 & 87.9 \\
        Duck        & 19.6 & 65.0 & 63.3 & 77.7 & 70.7 \\
        Eggbox      & 22.0 & 50.0 & 42.9 & 72.2 & 58.7 \\
        Glue        & 38.5 & 49.0 & 21.2 & 76.7 & 66.9 \\
        Hole p. & 22.1 & 39.0 & 42.1 & 91.4 & 90.6 \\
        \hline
        Average     & 24.9     & 40.8  & 52.6 & 78.0 & 75.1 \\
        \hline
    \end{tabular}
    \caption{Occlusion results with a single modality (left) and two modalities (right).}
    \label{tab:occlusion_color}
\end{table}

\section{Conclusion and future work}

In this work, we presented PointVoteNet, a method for 6 DoF object pose estimation using deep learning on point clouds. The developed algorithm is tested on two datasets. On the LINEMOD dataset, it outperforms other methods and achieves state-of-the-art performance. On the Occlusion dataset, the algorithm achieves comparable results with the current methods.

%PointVoteNet directly learns point to point correspondences between the object and scene. These correspondences are then fed into a pose voting scheme for pose estimation. By learning the object correspondences with deep learning, both local and global information can be used for the prediction. This allows for correct predictions even with heavy disturbances such as clutter and occlusion. The network is only trained using scene data, besides augmentations of the training data, no synthetic data is used.

The contribution is a novel framework for pose estimation using deep learning in 3D. PointNet was one of the first networks for training directly on unordered 3D data. Since then, a number of 3D point cloud networks with better performance have been developed \cite{qi2017pointnetplusplus,dgcnn,liu2019rscnn}. By directly replacing PointNet by these networks, our method can potentially improve.

% The current method can also be combined with 2D segmentation algorithms to increase performance. By using 2D segmentation, the points belonging to the object could be determined. From these points, the point to point correspondence can then be calculated. This will both decrease the algorithm runtime, and could possibly remove background points from the detection. 

\clearpage

\bibliographystyle{ieeetr}  
\bibliography{references}

\begin{thebibliography}{10}

\bibitem{brachmann2016uncertainty}
E.~Brachmann, F.~Michel, A.~Krull, M.~Ying~Yang, S.~Gumhold, {\em et~al.},
  ``Uncertainty-driven 6d pose estimation of objects and scenes from a single
  rgb image,'' in {\em IEEE Conference on Computer Vision and Pattern
  Recognition}, pp.~3364--3372, 2016.

\bibitem{rad2017bb8}
M.~Rad and V.~Lepetit, ``Bb8: A scalable, accurate, robust to partial occlusion
  method for predicting the 3d poses of challenging objects without using
  depth,'' in {\em IEEE International Conference on Computer Vision},
  pp.~3828--3836, 2017.

\bibitem{tekin2018real}
B.~Tekin, S.~N. Sinha, and P.~Fua, ``Real-time seamless single shot 6d object
  pose prediction,'' in {\em IEEE Conference on Computer Vision and Pattern
  Recognition}, pp.~292--301, 2018.

\bibitem{redmon2016you}
J.~Redmon, S.~Divvala, R.~Girshick, and A.~Farhadi, ``You only look once:
  Unified, real-time object detection,'' in {\em IEEE Conference on Computer
  Vision and Pattern Recognition}, pp.~779--788, 2016.

\bibitem{xiang2017posecnn}
Y.~Xiang, T.~Schmidt, V.~Narayanan, and D.~Fox, ``Posecnn: A convolutional
  neural network for 6d object pose estimation in cluttered scenes,'' {\em
  Robotics: Science and Systems}, 2018.

\bibitem{peng2019pvnet}
S.~Peng, Y.~Liu, Q.~Huang, X.~Zhou, and H.~Bao, ``Pvnet: Pixel-wise voting
  network for 6dof pose estimation,'' in {\em IEEE Conference on Computer
  Vision and Pattern Recognition}, pp.~4561--4570, 2019.

\bibitem{brachmann2014learning}
E.~Brachmann, A.~Krull, F.~Michel, S.~Gumhold, J.~Shotton, and C.~Rother,
  ``Learning 6d object pose estimation using 3d object coordinates,'' in {\em
  European conference on computer vision}, pp.~536--551, Springer, 2014.

\bibitem{kehl2017ssd}
W.~Kehl, F.~Manhardt, F.~Tombari, S.~Ilic, and N.~Navab, ``Ssd-6d: Making
  rgb-based 3d detection and 6d pose estimation great again,'' in {\em IEEE
  International Conference on Computer Vision}, pp.~1521--1529, 2017.

\bibitem{liu2016ssd}
W.~Liu, D.~Anguelov, D.~Erhan, C.~Szegedy, S.~Reed, C.-Y. Fu, and A.~C. Berg,
  ``Ssd: Single shot multibox detector,'' in {\em European conference on
  computer vision}, pp.~21--37, Springer, 2016.

\bibitem{wang2019densefusion}
C.~Wang, D.~Xu, Y.~Zhu, R.~Mart{\'\i}n-Mart{\'\i}n, C.~Lu, L.~Fei-Fei, and
  S.~Savarese, ``Densefusion: 6d object pose estimation by iterative dense
  fusion,'' in {\em IEEE Conference on Computer Vision and Pattern
  Recognition}, pp.~3343--3352, 2019.

\bibitem{qi2017pointnet}
C.~R. Qi, H.~Su, K.~Mo, and L.~J. Guibas, ``Pointnet: Deep learning on point
  sets for 3d classification and segmentation,'' in {\em IEEE Conference on
  Computer Vision and Pattern Recognition}, pp.~652--660, 2017.

\bibitem{rusu20113d}
R.~B. Rusu and S.~Cousins, ``3d is here: Point cloud library (pcl),'' in {\em
  IEEE international conference on robotics and automation}, pp.~1--4, IEEE,
  2011.

\bibitem{buch2017rotational}
A.~G. Buch, L.~Kiforenko, and D.~Kraft, ``Rotational subgroup voting and pose
  clustering for robust 3d object recognition,'' in {\em IEEE International
  Conference on Computer Vision}, pp.~4137--4145, 2017.

\bibitem{besl1992method}
P.~Besl and N.~D. McKay, ``A method for registration of 3-d shapes,'' {\em IEEE
  Transactions on Pattern Analysis and Machine Intelligence}, vol.~14, no.~2,
  pp.~239--256, 1992.

\bibitem{hinterstoisser2012model}
S.~Hinterstoisser, V.~Lepetit, S.~Ilic, S.~Holzer, G.~Bradski, K.~Konolige, and
  N.~Navab, ``Model based training, detection and pose estimation of
  texture-less 3d objects in heavily cluttered scenes,'' in {\em Asian
  conference on computer vision}, pp.~548--562, Springer, 2012.

\bibitem{kingma2014adam}
D.~P. Kingma and J.~Ba, ``Adam: A method for stochastic optimization,'' {\em
  arXiv preprint arXiv:1412.6980}, 2014.

\bibitem{qi2017pointnetplusplus}
C.~R. Qi, L.~Yi, H.~Su, and L.~J. Guibas, ``Pointnet++: Deep hierarchical
  feature learning on point sets in a metric space,'' in {\em Advances in
  neural information processing systems}, pp.~5099--5108, 2017.

\bibitem{dgcnn}
Y.~Wang, Y.~Sun, Z.~Liu, S.~E. Sarma, M.~M. Bronstein, and J.~M. Solomon,
  ``Dynamic graph cnn for learning on point clouds,'' {\em ACM Transactions on
  Graphics}, 2019.

\bibitem{liu2019rscnn}
Y.~Liu, B.~Fan, S.~Xiang, and C.~Pan, ``Relation-shape convolutional neural
  network for point cloud analysis,'' in {\em IEEE Conference on Computer
  Vision and Pattern Recognition}, pp.~8895--8904, 2019.

\end{thebibliography}

\end{document}